\documentclass[10pt,twocolumn,letterpaper]{article}

\usepackage{cvpr}
\usepackage{times}
\usepackage{epsfig}
\usepackage{graphicx}
\usepackage{amsmath}
\usepackage{amssymb}
\usepackage{multirow}

\usepackage[breaklinks=true,bookmarks=false]{hyperref}

\cvprfinalcopy % *** Uncomment this line for the final submission

\def\cvprPaperID{0012} % *** Enter the CVPR Paper ID here

\ifcvprfinal\fi
\begin{document}

%%%%%%%%% TITLE
\title{Deep Robust Single Image Depth Estimation Neural Network \\ Using Scene Understanding}

\author{Haoyu Ren, Mostafa El-khamy, Jungwon Lee\\
SOC R\&D, Samsung Semiconductor Inc.\\
9868 Scranton Road, San Diego, CA, US\\
{\tt\small \{haoyu.ren, mostafa.e, jungwon2.lee\}@samsung.com}
%\and
%\\
}

\maketitle
%\thispagestyle{empty}

%%%%%%%%% ABSTRACT
\begin{abstract}
\noindent{}Single image depth estimation (SIDE) plays a crucial role in 3D computer vision. In this paper, we propose a two-stage robust SIDE framework that can perform blind SIDE for both indoor and outdoor scenes.  At the first stage, the scene understanding module will categorize the RGB image into different depth-ranges. We introduce two different scene understanding modules based on scene classification and coarse depth estimation respectively. At the second stage, SIDE networks trained by the images of specific depth-range are applied to obtain an accurate depth map. In order to improve the accuracy, we further design a multi-task encoding-decoding SIDE network DS-SIDENet based on depthwise separable convolutions. DS-SIDENet is optimized to minimize both depth classification and  depth regression losses. This improves the accuracy compared to a single-task SIDE network. Experimental results demonstrate that training DS-SIDENet on an individual dataset such as NYU achieves competitive performance to the state-of-art methods with much better efficiency.  Ours proposed robust SIDE framework also shows good performance for the ScanNet indoor images and KITTI outdoor images simultaneously. It achieves the top performance compared to the Robust Vision Challenge (ROB) 2018 submissions.
\end{abstract}

%%%%%%%%% BODY TEXT
\section{Introduction}
\noindent{}Single image depth estimation (SIDE) is a key feature of understanding the geometric structure of the scene. In particular, the depth map can be used to infer the 3D structure, which is the basic element of many topics in 3D vision, such as image reconstruction, image rendering, and shallow depth of the field. SIDE is an ill-posed problem since a single 2D image may be produced from an infinite number of distinct 3D scenes. To overcome this ambiguity, typical methods focus on exploiting statistically meaningful features, such as perspective and texture information, object locations, and occlusions. Recently, with the prosperity of deep convolutional neural networks (CNNs), many deep learning-based methods have achieved significant performance improvement. 

%These methods address the SIDE problem as learning a CNN mapping the input RGB image to the continuous depth map. 

This work aims to address two main issues of deep CNNs for SIDE. Most of current SIDE methods train different networks for individual datasets, which makes models be specific to certain domains. As a result, the large differences between different datasets such as indoor and outdoor patterns limit the generalization ability of the SIDE network. A model achieving considerable performance on one dataset will perform poorly on the other one. Hence, the first issue we aim to address is how to design a framework for blind SIDE, where both indoor images and outdoor images are fed into same SIDE network. Existing SIDE networks utilize very deep architecture such as ResNet-101 \cite{fu2018deep} or ResNext-101 \cite{li2018deep} as backbone. The sizes of these networks are very large. In addition, some of existing SIDE networks such as DORN \cite{fu2018deep} contain fully connected layers. So a sliding-window strategy is required during the prediction to make the testing image size same as training. As a result, the efficiency and flexibility of such networks is not very good. Hence, the second issue we aim to address is designing an efficient fully-convolutional SIDE network achieving high accuracy as well. 

\begin{figure}
\centering
\includegraphics[width=8cm]{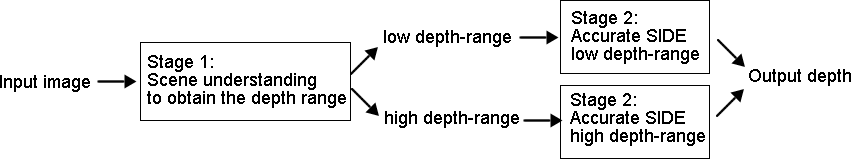}
\caption{Our proposed two-stage robust SIDE framework.}
\label{fig:Robust-SIDEs}
\end{figure}

In this paper, we propose an effective two-stage framework for robust single image depth estimation, as shown in Fig. \ref{fig:Robust-SIDEs}. Our framework starts by a scene understanding module, where the input RGB image is categorized into low depth-range or high depth-range. Then we apply different SIDE networks trained by low depth-range RGB images and high depth-range images respectively, to estimate an accurate depth map. Since we do not have any prior knowledge before the testing, our framework is blind SIDE. We show that the scene understanding module can be implemented by either a scene classification network, or a coarse single image depth estimation network. Both of them work well for the robust SIDE task of  Robust Vision Challenge (ROB) 2018 \cite{rob2018link}, where the testing images are randomly sampled from ScanNet \cite{dai2017scannet} and KITTI\cite{geiger2013vision} datasets.

Next, we present an efficient fully-convolutional neural network DS-SIDENet for single image depth estimation.  DS-SIDENet is a multi-task network following the encoding-decoding architecture \cite{laina2016deeper}. The encoding part consists of several depthwise separable (DS) convolutional layers to extract discriminative features from the input image. The decoding part consists of two branches corresponding to the depth classification task and the depth regression task. In the depth classification branch, the network generates pixel-wise classification labels of the quantized depth. In the depth regression branch, a continuous depth map is directly estimated. During the training, we optimize the loss functions considering these two branches at the same time, while only the depth classification result is utilized during the predication. This improves the discriminative ability of the network, without adding computational cost during depth estimation. Our multi-task network is different from existing multi-task networks \cite{eigen2015predicting} which considers depth estimation and semantic segmentation together. The two tasks in DS-SIDENet can be considered as different formulations of same depth estimation problem, instead of different problems in \cite{eigen2015predicting}. We show that DS-SIDENet can achieve better performance than existing SIDE networks in NYU-Depth-v2 \cite{eigen2014depth} dataset. Our proposed robust SIDE based on DS-SIDENet achieves the 1st rank compared to ROB 2018 submissions.  

The contributions of our work can be summarized as:

- We propose a two-stage framework effective for robust SIDE based on scene understanding. Two different scene understanding modules, scene classification and coarse depth estimation, are discussed

- We design an efficient SIDE network based on depthwise separable convolution. It achieves considerable accuracy and efficiency concurrently

- We show that doing multi-task learning in SIDE by considering depth classification and depth regression together is helpful to improve the overall accuracy

%The rest of this paper is organized as follows. Section~2 gives a brief survey of the related works on SIDE. Section~3 describes the proposed DS-SIDENet. Section~4 shows the details of the robust SIDE framework. In Section~5, the experimental results of NYU, ScanNet, and KITTI are given, as well as comparison to the state-of-art methods. Conclusions are made in the last section.

\section{Related work}
\noindent{}Previous approaches for depth estimation from single images can be categorized into two main groups, methods operating on hand-crafted features, and methods adopting deep neural networks. Earlier works addressing the depth estimation task belong to the first category. Hoiem et al. \cite{hoiem2005automatic} introduced photo pop-up, a fully automatic method for creating a basic 3D model from a single photograph. Karsch et al. \cite{karsch2014depth} developed Depth Transfer, a non-parametric approach where the depth of an input image is reconstructed by transferring the depth of multiple similar images with warping and optimizing procedures. Ladicky \cite{ladicky2014pulling} demonstrated the benefit of combining semantic object labels with depth features. Saxena et al. \cite{saxena20083} introduced a multi-scale conditional random field (CRF) to extract multi-scale context information for depth estimation.

More recent approaches for depth estimation are based on convolutional neural networks. In the pioneer work \cite{eigen2014depth}, Eigen et al. introduced a coarse-to-fine network, and utilized the scale-invariant loss to improve the accuracy of the estimated depth map. This work is further extended in \cite{eigen2015predicting}, where the depth estimation, surface normal estimation, and semantic segmentation are integrated into one unified network. Li et al. \cite{li2017two} considered a loss function with components in both the depth domain and the gradient of depth domain. Fu. et al. \cite{fu2018deep} introduced a spacing-increasing discretization strategy to discretize depth and re-casted depth network learning as an ordinal regression problem. Their method achieves significant accuracy improvement compared to previous methods. 

% There are a few works who consider combining the CNN and CRF together. In conventional CRF-based depth estimation, the observation is the RGB image. A feasible way is to use the CNN feature maps instead of the RGB image. Similar structure has already been utilized in semantic segmentation. For instance, in \cite{mousavian2016joint}, Mousavian et al. proposed to combine the depth estimation and semantic segmentation together. First, a CNN is utilized to extract the feature maps of the depth and semantic labels at the same time. Then, these feature maps are fused as the input of a CRF. Xu et al. \cite{xu2017multi} propose a cascade structure for CNN-CRF depth estimation. They use the side output of CNN layers as the input to CRF, and perform depth estimation at a certain scale, and then refine the obtained estimates in the subsequent level. Two CRF architectures, multiscale CRF (one CRF use all feature maps in the potential function), and cascade CRF (multi CRFs, where the depth of the previous CRF is the input of the next CRF) are proposed. In addition, they implement the above CRFs as neural network layers. As a result, the depth estimation can be achieved by an end-to-end neural network. The training loss is the L$2$ difference between the estimated depth and the ground-truth depth.

Most of above works use some backbones with fully connected layer. This will increase the model complexity and computation cost. In addition, the input image size is restricted during the testing. To solve this problem, a fully convolutional network  was proposed by Laina et al. \cite{laina2016deeper}. A revised version of this work is introduced in \cite{ma2017sparse}, where randomly sampled sparse depth is adopted together with the RGB image to predict a dense depth map. Chen et al. \cite{chen2016single} used a fully-convolutional network to predict the relative depth map, e.g., a relationship between the depths of any two pixels. Kuznietsov et al. \cite{kuznietsov2017semi} utilized semi-supervised learning for SIDE, where supervised learning on sparse measurements is complemented with unsupervised learning of the left-right consistency in stereo images.  Cheng et al. \cite{cheng2018depth} proposed a spatial propagation network to learn the affinity matrix and showed its effectiveness to improve the performance of existing SIDE networks. All of the above CNNs are constructed based on some computationally expensive architecture. In contrast, our DS-SIDENet is constructed by depthwise separable convolution \cite{ren2018dn}, which is more efficient during the prediction.

% Recently, the state-of-art single image depth estimation is achieved by DORN \cite{fu2018deep}. In \cite{li2017single}, Li et al. reformulated this problem as a classification problem, and used a network with dilated convolutions to classify depth into several discrete segments. The authors introduce a quantization strategy to discretize the depth and recast the depth estimation problem as an ordinal regression problem. By training the network using an ordinary regression loss, DORN achieves much higher accuracy with faster convergence. A multiscale network structure which avoids unnecessary spatial pooling and captures multiscale information in parallel is further utilized.

\section{Robust single image depth estimation}
\noindent{}Blind depth estimation is very important in real scenario. A good SIDE system should consequently perform well on a variety of datasets with different characteristics, e.g., for both the indoor scenario and the outdoor scenario. One straightforward solution is to mix the training data of indoor and outdoor images to train one network, which is utilized by many of the ROB submissions. For example, Li et al. \cite{li2018deep} resized the KITTI and ScanNet images to $320\times256$, and utilized a channel-wise attention mechanism to adaptively select the discriminative channels of features for indoor and outdoor scenarios. But this will decrease the accuracy compared to training different networks for different datasets. 

\begin{figure}
\centering
\includegraphics[width=8cm]{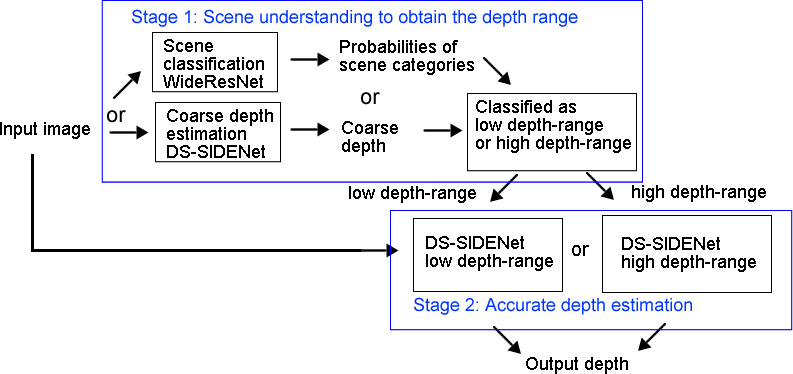}
\caption{Robust SIDE based on two different scene understanding modules.}
\label{fig:Robust-SIDE}
\end{figure}

The major difficulty of blind SIDE is the contextual difference between scenes. Different scene types tend to have different depth-ranges. For example, outdoor scenes tend to have a larger depth-range than indoor scenes. So we propose to use scene understanding to capture the depth-range, in order to support the following depth estimation. We introduce a two-stage framework as illustrated in Fig. \ref{fig:Robust-SIDE}. The first stage is a scene understanding module, which classifies the input RGB image into  low depth-range category (all objects are close to camera) or high depth-range category (including objects far away, e.g., distance higher than threshold). Then we apply different SIDE networks for the low depth-range images and high depth-range images respectively to get an accurate depth map. In this section, we give two  different solutions for the scene understanding stage.

%\subsection{Using scene classification as scene understanding}

%There are two disadvantages of such dataset fusion. Firstly, the accuracy will be worse compared to training the network in single dataset. Secondly, due to the using of fully-connected layer, the testing image size has to be the same as the training image size. This reduces the flexibility of the network.

Our first solution is using the scene classification methods to get the probability of the scenes, and obtaining the depth-range from these probabilities. Assume we have   scene categories with `low depth-range' or `high depth-range' labels based on their context, given the estimated probabilities of these scenes from a scene classification network, the depth-range can be calculated by a `majority voting' from the top-$K$ scenes with highest probability. For example, the input image is classified as `low depth-range' if there are more than $K/2$ `low depth-range' votes from the top-$K$ scene categories. Other methods can also be used to decide the depth-range such as weighted voting, or the expected depth-range based on the predicted scene probabilities and the average depth-range of each scene. In this paper, we show that even the simple majority voting strategy is good in the ROB challenge. In our implementation, we use the WideResNet-18  trained on the Places-365 dataset \cite{zhou2017places} as the scene classification network. This dataset consists of 365 labeled scene  categories, such as sky, arena, iceberg, etc. Given an input image, we use the top-15 ($K=15$) predicted scene categories to vote for depth-range. 

%More specifically, the input image is classified as `low depth-range' if there are more than 7 `low depth-range' votes from the top-15 scene categories. 

%The pre-trained Places-365 model is directly used, without any fine-tuning in our evaluation.

%\subsection{Using coarse depth estimation for robust SIDE}
Our second solution is training a coarse depth estimation network by mixing images from different depth-ranges together. Then the depth-range can be inferred from the statistics of the estimated coarse depth map. We calculate the maximum depth $d_{max}$ of the whole estimated depth map, and compare it to a threshold $\sigma$. If $d_{max}>\sigma$, the RGB input is categorized as of high depth-range, otherwise it will be categorized as of low depth-range. To decide $\sigma$, we calculate the maximum estimated depth statistic using the validation images of the ROB 2018 challenge \cite{rob2018link}, from which we select $\sigma$ as $5.89$. In our implementation, the coarse depth-range estimation network has the same architecture as that of  the low depth-range DS-SIDENet given in Table \ref{table:encoding} and Table \ref{table:decoding}.

 %Since the low depth-range images and high depth-range images have totally different characteristics, this coarse depth estimation network will not be very accurate when calculating pixel-wise error

After obtaining the depth-range, we apply one of the two DS-SIDENets optimized for the low depth-range images and high depth-range images to generate an accurate depth map. We also investigated the `coarse-to-fine' architecture proposed in \cite{eigen2014depth}\cite{ma2017sparse}, by utilizing an RGBD network which takes the estimated depth from the coarse depth estimation as an additional feature with the RGB channels . However, due to the inaccurate depth estimates from the coarse network, no significant performance gain was observed. It will be difficult for RGBD depth refinement with an inaccurate  sparse depth features.
%Using  inaccurate input in the RGBD refinement network will decrease the performance.  The potential reason is that the estimated coarse depth  is not very accurate (as shown in Table \ref{table:ScanNet accuracy} and \ref{table:KITTI accuracy}).

\section{DS-SIDENet}
\begin{figure*}
\centering
\includegraphics[width=16cm]{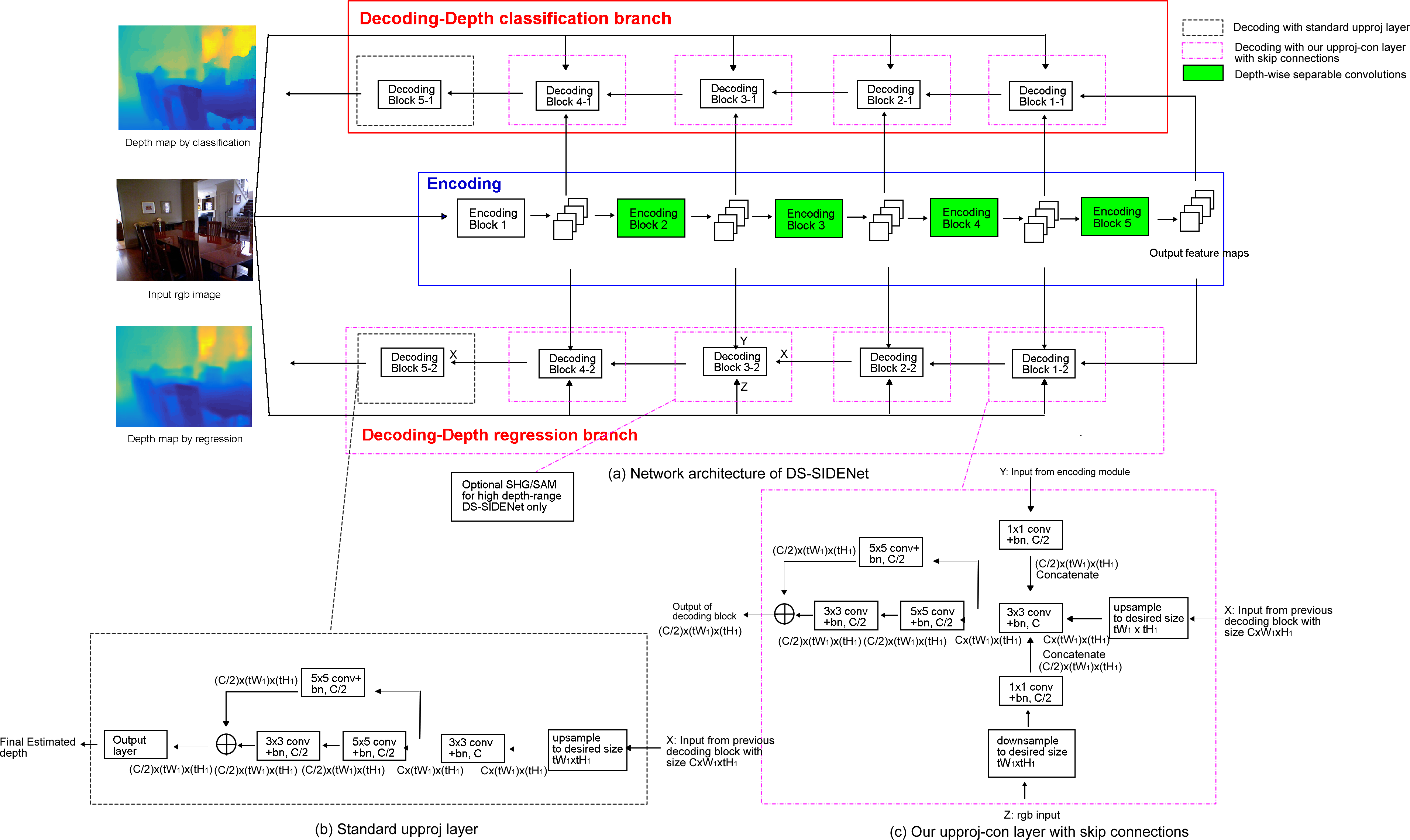}
\caption{ (a) Network architecture of DS-SIDENet. (b) The standard upproj layer in \cite{laina2016deeper}. The `output layer' in classification branch is different from the one in regression branch. (c) Our upproj-con layer with skip connections. The `X', `Y' and `Z' correspond to the three input feature maps of this block.  `t' is the scale difference between the input feature map and output feature map. `${C}\times{}{W}_1\times{}{H}_1$' is the size of input feature maps of the decoding block. }
\label{fig:DW-SIDE}
\end{figure*}

\noindent{}We propose a single image depth estimation network DS-SIDENet based on depthwise separable convolutions and encoder-decoder architecture  \cite{laina2016deeper}. As shown in Fig. \ref{fig:DW-SIDE}, our network consists of two parts, the encoding part, and a two-branch decoding part. 

\subsection{Encoding}
\noindent{}The encoding part contains several  depthwise separable convolutional layers \cite{ren2018dn} to extract the discriminative features from the input image, as given in  Table \ref{table:encoding}. Each depthwise separable convolutional layer consists of a depthwise  convolutional layer and a point-wise convolutional layers. The low depth-range images have different characteristics with the high depth-range images, e.g., indoor and outdoor images have different context. The encoding part of low depth-range network (e.g., trained in NYU) is slightly different from the high depth-range network (e.g., trained in KITTI). We downsample the output feature map size of the high depth-range encoding module by $\times4$, and the low depth-range encoding module by $\times32$. 

%Similar to U-Net \cite{ronneberger2015u}, the intermediate feature maps from the encoding blocks are connected to the decoding blocks. We also add a skip connection from the RGB input to the decoding blocks to capture the global context.

\begin{table*}
\caption{Network architecture of the encoding part of DS-SIDENet. The `Block1' to `Block5' correspond to the encoding blocks displayed in Fig. \ref{fig:DW-SIDE}(a).`s' means stride, `d' means dilation rate, `dw' means depthwise separable convolutions. `convbn' layer includes a convolutional layer, followed by a batch normalization layer.}
\scriptsize
\renewcommand{\arraystretch}{1.25}
\centering
\begin{tabular}{|c|c|c|c|c|}
\hline
\multirow{2}{*}{layer} & \multicolumn{2}{|c|}{low depth-range network} & \multicolumn{2}{|c|}{high depth-range network}  \\ \cline{2-5}
& layer setting & output dimension & layer setting & output dimension \\ \hline
input& & $3\times{}W\times{}H$ & & $3\times{}W\times{}H$ \\ \hline
Block 1  & $\begin{aligned}convbn,3\times3,32,s2,d1 \\ convbn,3\times3,64,s1,d1 \end{aligned} $ & $64\times{}\frac{1}{2}W\times{}\frac{1}{2}H $  &  $\begin{aligned}convbn,3\times3,32,s2,d1 \\ convbn,3\times3,64,s1,d1 \end{aligned} $ & $64\times{}\frac{1}{2}W\times\frac{1}{2}H$\\\hline
Block 2  & $\begin{aligned}\quad [convbn,3\times3,64,s1,d1,dw]\times 3\\ convbn,3\times3,128,s2,d1,dw \quad \end{aligned} $ & $128\times{}\frac{1}{4}W\times{}\frac{1}{4}H $  &  $\begin{aligned}\quad [convbn,3\times3,64,s1,d1,dw]\times 3\\ convbn,3\times3,128,s2,d1,dw \quad \end{aligned} $ & $128\times{}\frac{1}{4}W\times\frac{1}{4}H$\\\hline
Block 3 & $\begin{aligned}\quad [convbn,3\times3,128,s1,d1,dw]\times 3\\ convbn,3\times3,256,s2,d1,dw \quad \end{aligned} $ & $256\times{}\frac{1}{8}W\times{}\frac{1}{8}H $  &  $\begin{aligned}\quad [convbn,3\times3,128,s1,d1,dw]\times 3\\ convbn,3\times3,256,s1,d1,dw \quad \end{aligned} $ & $256\times{}\frac{1}{4}W\times\frac{1}{4}H$\\ \hline
Block 4 & $\begin{aligned}\quad [convbn,3\times3,32,s1,d2,dw]\times 12\\ convbn,3\times3,256,s2,d1,dw \quad \end{aligned} $ & $256\times{}\frac{1}{16}W\times{}\frac{1}{16}H$  &  $\begin{aligned}\quad [convbn,3\times3,32,s1,d2,dw]\times 12\\ convbn,3\times3,256,s1,d1,dw \quad \end{aligned} $ & $256\times{}\frac{1}{4}W\times\frac{1}{4}H$\\\hline
Block 5 & $\begin{aligned}convbn,3\times3,256,s2,d1,dw \\ convbn,1\times1,512,s1,d1 \end{aligned} $ &  $512\times{}\frac{1}{32}W\times\frac{1}{32}H$ &  $\begin{aligned}convbn,3\times3,256,s1,d1,dw \\ convbn,1\times1,128,s1,d1 \end{aligned} $  & $128\times{}\frac{1}{4}W\times\frac{1}{4}H$\\ \hline
\end{tabular}
\label{table:encoding}
\end{table*}

\subsection{Decoding}
\noindent{}Since the output feature maps of the encoding part are downsampled, the decoding part requires several upsampling modules to retrieve the input feature map size. A baseline solution is the `upproj' layer proposed in \cite{laina2016deeper}, as given in Fig. \ref{fig:DW-SIDE}(b), which achieves better performance compared to upsampling by simple deconvolution layer. Inspired by the U-Net architecture \cite{ronneberger2015u}, we design an improved upsampling layer with additional skip connections from the encoding part and RGB input, denoted as `upproj-con' layer. As shown in Fig. \ref{fig:DW-SIDE}(c), assume the input feature map size from the previous decoding layer is $C\times{}W_1\times{}H_1$, the output feature map size of `upproj-con' layer will be $\frac{C}{2}\times{}tW_1\times{}tH_1$, where $t$ is the scale difference between the input feature map and output feature map.

The decoding part consists of two branches, the depth classification branch and the depth regression branch. These two branches have mostly the same network architecture, but they are optimized with different loss functions. Each of these two branches consists of several `upproj-con' decoding blocks and one additional `upproj' block placed before obtaining the final depth output, as given in Fig. \ref{fig:DW-SIDE}(a) and Table \ref{table:decoding}. For the low depth-range network, we set all $t=2$. For the high depth-range network, we set some of these $t$ to 1 since the output of the encoding part of high depth-range network is only downsampled 4 times. 
In addition, we replace the third  upsampling module by stacked network modules such as stacked hourglass (SHG) since it shows superior performance in the outdoor disparity estimation task \cite{chang2018pyramid}. We further adopt the stacked atrous multi-scale (SAM) module proposed in \cite{du2019amnet} instead of standard SHG, which replaces the downsampling and upsampling  in SHG by atrous convolutions. An ablation study of using different decoding blocks is given in section 5.3.2.

\begin{table*}
\caption{Network architecture of each of the decoding branch in DS-SIDENet. `Block i' corresponds to the `Decoding block i-1' and `Decoding block i-2' in Fig. \ref{fig:DW-SIDE}(a), i=1,2,...,5. `SAM' has exact same architecture as the one in \cite{du2019amnet}, where 3 AM modules are stacked.}
\scriptsize
\renewcommand{\arraystretch}{1.25}
\centering
\begin{tabular}{|c|c|c|c|c|}
\hline
\multirow{2}{*}{layer} & \multicolumn{2}{|c|}{low depth-range network} & \multicolumn{2}{|c|}{high depth-range network}  \\ \cline{2-5}
& layer setting & output dimension & layer setting & output dimension \\ \hline
Input feature maps& & $512\times{}\frac{1}{32}W\times\frac{1}{32}H$ & & $128\times{}\frac{1}{4}W\times\frac{1}{4}H$ \\\hline
Block 1 & upproj-con,$t=2$ & $256\times{}\frac{1}{16}W\times{}\frac{1}{16}H $  &  upproj-con,$t=1$ & $64\times{}\frac{1}{4}W\times\frac{1}{4}H$\\ \hline
Block 2 & upproj-con,$t=2$ & $128\times{}\frac{1}{8}W\times{}\frac{1}{8}H $  &  upproj-con,$t=1$ & $32\times{}\frac{1}{4}W\times\frac{1}{4}H$\\ \hline
Block 3 & upproj-con,$t=2$ & $64\times{}\frac{1}{4}W\times{}\frac{1}{4}H$  &  SAM & $32\times{}\frac{1}{4}W\times\frac{1}{4}H$\\\hline
Block 4 & upproj-con,$t=2$ & $32\times{}\frac{1}{2}W\times{}\frac{1}{2}H $  &  upproj-con,$t=2$ & $16\times{}\frac{1}{2}W\times{}\frac{1}{2}H $ \\ \hline
Block 5 & upproj,$t=2$& $1\times{}W\times{}H $ & upproj,$t=2$&  $1\times{}W\times{}H $ \\ \hline
\end{tabular}
\label{table:decoding}
\end{table*}

\subsection{Multi-task loss function}
\noindent{}In the classification branch, the network performs depth classification after quantizing the continuous depth into several bins. We follow the quantization scheme proposed in DORN \cite{fu2018deep}, where depth-range $[\alpha, \beta]$ is uniformly quantized into $K$ bins in the log scale. As shown in Eq. \ref{eq:quantize}, the input continuous depth $x$ is quantized to discrete value $b$, $K$ is the number of bins, and $q$ is the width of each bin.

\begin{equation}
\begin{aligned}
b = round(log_{10}(x)-log_{10}(\alpha))/q) \\ 
q=(log_{10}(\beta)-log_{10}(\alpha))/K
\end{aligned}
\label{eq:quantize}
\end{equation} 

When calculating the loss function of the classification branch, we use a soft classification loss similar to disparity estimation \cite{chang2018pyramid}, where the expected depth is the weighted sum of the quantized depth probabilities and the quantized depth values, and is the predicted  depth for each pixel location. Let $p_i^j$ represents the probability of pixel $i$ being quantized depth $j$, and $D$ is the maximum quantized depth $b_{max}$ in Eq. \ref{eq:quantize} when $x=\beta$, the expected quantized depth $d_i$ of pixel $i$ is calculated as 
$d_i=\sum_{j=1}^D{j\times p_i^j}$. A smooth $L_1$ loss is utilized to measure the difference $y=d_i-d_{i,qt}^*$ between the predicted quantized depth $d_i$ and the ground truth quantized depth $d_{i,qt}^*$, as given in Eq. \ref{eq:smoothl1}. 

%In section 5.3.1, we show that such classification loss function is better than existing depth classification losses. 

%\begin{equation}
%d_i=\sum_{j=1}^D{j\times p_i^j}.
%\label{eq:regression}
%\end{equation} 

\begin{equation}
 smooth_{L_1}(y)= \left\{
\begin{aligned}
0.5y^2 \quad \quad if |y|<1 \\
|y|-0.5 \quad otherwise
\end{aligned}
\right.
\label{eq:smoothl1}
\end{equation} 

In the regression branch, the network directly regresses the output feature maps to a continuous disparity map. We use the $L_1$ loss between the output and ground-truth depth to measure the regression accuracy. Both the classification branch and the regression branch are optimized simultaneously using multi-task learning, where the final loss function $L$ is a linear combination of the above two loss functions, as shown in Eq. \ref{eq:l1loss}, where $d_{i,qt}^*$ is the quantized depth ground truth, and $d_{i}^*$ is the continuous depth ground truth. The weights $w_1$ and $w_2$ are set to 1 empirically.
 
\begin{equation}
L = \sum_i{w_1\cdot{}smooth_{L_1}(d_{cls,i}, d_{i,qt}^*)+w_2\cdot{} L_{1}(d_{reg,i}, d_i^*)}.
\label{eq:l1loss}
\end{equation} 

In section 5.3.1, we show that the depth maps from either the classification branch or the regression branch of DS-SIDENet is better than the outputs from the single task SIDE network. We also observe that the classification output is more accurate than regression output. We only use the classification output during the testing, so that the computational cost will not increase.

\section{Experiments}
\subsection{Datasets and network training}
\noindent{}We first utilize NYU dataset to show the effectiveness of the proposed DS-SIDENet. The NYU-Depth-v2 \cite{eigen2014depth} dataset consists of RGB and depth images collected from 464 different indoor scenes. We use the official split of data as \cite{eigen2014depth}, where 249 scenes are used for training and the rest 215 for testing. Since all NYU images are indoor images with relatively small depth, we train a low depth-range DS-SIDENet on this dataset.

Next, we follow the robust single image depth estimation challenge in ROB 2018 \cite{rob2018link}, where the same depth estimation system is evaluated on both the ScanNet indoor images and the KITTI street-view images. The ScanNet \cite{dai2017scannet} is a large-scale RGB-D dataset for indoor scene reconstruction, which contains 2.5 million RGB images in 1,513 scenes. Original RGB images are captured at a resolution of $1296\times{}968$ and depth at $640\times{}480$. The KITTI single image depth prediction dataset \cite{geiger2013vision} is a real-world dataset with street views from a driving car. It contains about 46,000 training stereo image pairs with sparse ground-truth depth. KITTI image size is around $1240\times{}370$. We use the official ROB training/validation/testing split for these two datasets.

Our robust SIDE framework consists of three modules, the scene understanding, low depth-range DS-SIDENet, and high depth-range DS-SIDENet. We resize ScanNet images to $320\times240$ to train the low depth-range DS-SIDENet, and zero pad KITTI images to $1248\times384$ to train the high depth-range DS-SIDENet. As mentioned in Section 3, the scene understanding can be achieved by either a scene classification network, or a coarse depth estimation network. For scene classification, we utilize the official model of WideResNet-18 trained on the Places-365 dataset, and resize the input RGB image to $224\times224$ during the testing to predict the scene probabilities. For coarse depth estimation, we mix the training images from ScanNet and KITTI to train a DS-SIDENet. The ScanNet training images are downsampled to $320\times{}240$. The KITTI training images are zero padded to $1248\times384$ and downsampled to $832\times{}256$. Then random $320\times{}240$ patches are cropped to keep the same size as ScanNet. 

We utilize multiple evaluation metrics to evaluate the performance of DS-SIDENet, including mean relative absolute error (REL), mean relative squared error (sqREL), root mean square error (RMSE), inverse root mean square error (iRMSE), scale invariant logarithmic error (SILog), mean average error (MAE), and $\delta$ threshold  ($\delta_i = \delta < 1.25^i$). More details can be found in \cite{eigen2014depth}\cite{cheng2018depth}. 

%- Mean relative absolute error (REL) $\frac{1}{N}\sum_{i=1}^N\frac{|d_i-d_i^*|}{d_i^*}$

%- Mean relative squared error (sqREL) $\frac{1}{N}\sum_{i=1}^N\frac{|d_i-d_i^*|^2}{d_i^*}$

%- Root mean squre error (RMSE)$\sqrt{\frac{1}{N}\sum_{i=1}^N{|d_i-d_i^*|}^2}$

%- Inverse root mean square error (iRMSE)
%$\sqrt{\frac{1}{N}\sum_{i=1}^N{|\frac{1}{d_i}-\frac{1}{d_i^*}|}^2}$

%- Scale invariant logarithmic error (SILog)
%$\frac{1}{N}\sum_{i=1}^N{log{d_i}-log{d_i^*}}+\frac{1}{|S_i|}\sum_{j\in{}S_i}{logd_j^*-logd_j}$

%where $d_i$ is the estimated depth, $d_i^*$ is the ground truth depth of pixel $i$, $N$ is the number of valid pixels in the test images, $S_i$ is a set of pixels that belong to the same images with pixel $i$.

\subsection{Comparison to existing methods}
\begin{figure}
\centering
\includegraphics[width=8.5cm]{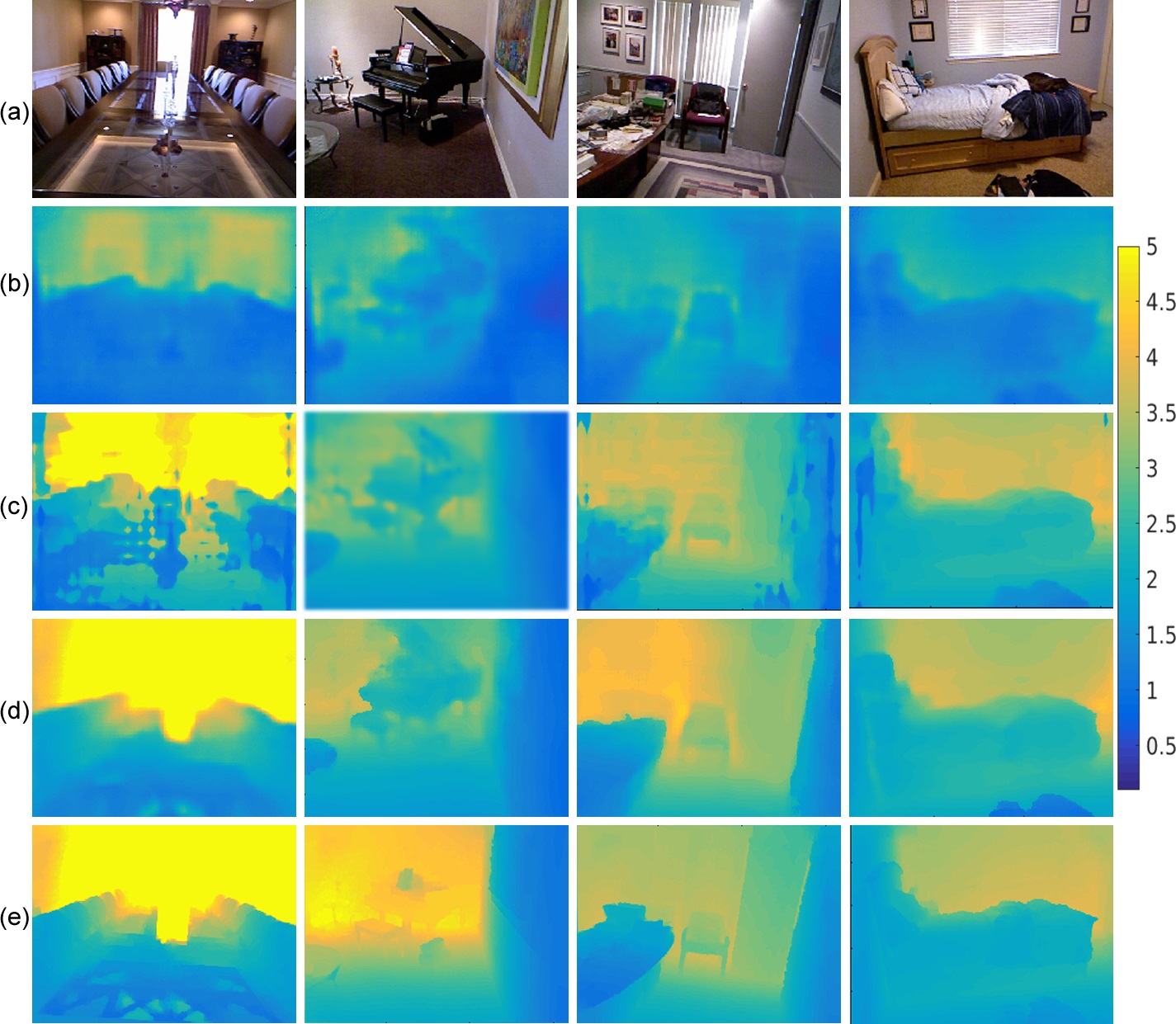}
\caption{Example outputs of DS-SIDENet in NYU dataset, as well as state-of-art SIDE methods. (a) RGB input (b) Laina \cite{laina2016deeper} (c) DORN \cite{fu2018deep} (d) Ours DS-SIDENet (e) Ground-truth.}
\label{fig:NYU}
\end{figure}

\noindent{}First we use the NYU dataset to evaluate our DS-SIDENet. In Table \ref{table:NYU accuracy}, we give the comparison of our DS-SIDENet to some commonly-used SIDE methods. It can be seen that our DS-SIDENet achieves better accuracy than the state-of-art SIDE method DORN \cite{fu2018deep}. In addition, our DS-SIDENet takes only 150ms to predict a $640\times480$ image in single Tesla V100 GPU, which is much faster than  600 ms of DORN, and 1 second of MS-CRF. This signifies the advantage of depthwise separable convolutions. Some example outputs are given in Fig. \ref{fig:NYU}, where better estimated depth maps are generated by DS-SIDENet.

\begin{table}
\caption{Depth estimation accuracy of different SIDE networks in NYU dataset. }
\small
\renewcommand{\arraystretch}{1.0}
\centering
\begin{tabular}{|c|c|c|c|c|c|}
\hline
Method & REL & RMSE &$\delta_1$ & $\delta_2$ & $\delta_3$  \\ \hline
Eigen \cite{eigen2015predicting} & 0.158 & 0.641 & 0.769 & 0.950 & 0.988 \\
Chakrabarti \cite{chakrabarti2016depth} & 0.149 & 0.620 & 0.806 & 0.950 & 0.988 \\
Laina \cite{laina2016deeper} & 0.127 & 0.573 & 0.629 & 0.899 & 0.971 \\
Li \cite{li2017two} & 0.152 & 0.611 & 0.789 & 0.955 & 0.988 \\
MS-CRF \cite{xu2017multi} & 0.121 & 0.586 & 0.811 & 0.954 & 0.987 \\
DORN \cite{fu2018deep} & 0.115 & 0.509 & 0.828 & 0.965 & 0.992 \\ \hline
DS-SIDENet & \textbf{0.113} & \textbf{0.501} & \textbf{0.833} & \textbf{0.968} & \textbf{0.993} \\ \hline

\end{tabular}
\label{table:NYU accuracy}
\end{table}

\begin{figure}
\centering
\includegraphics[width=8.5cm]{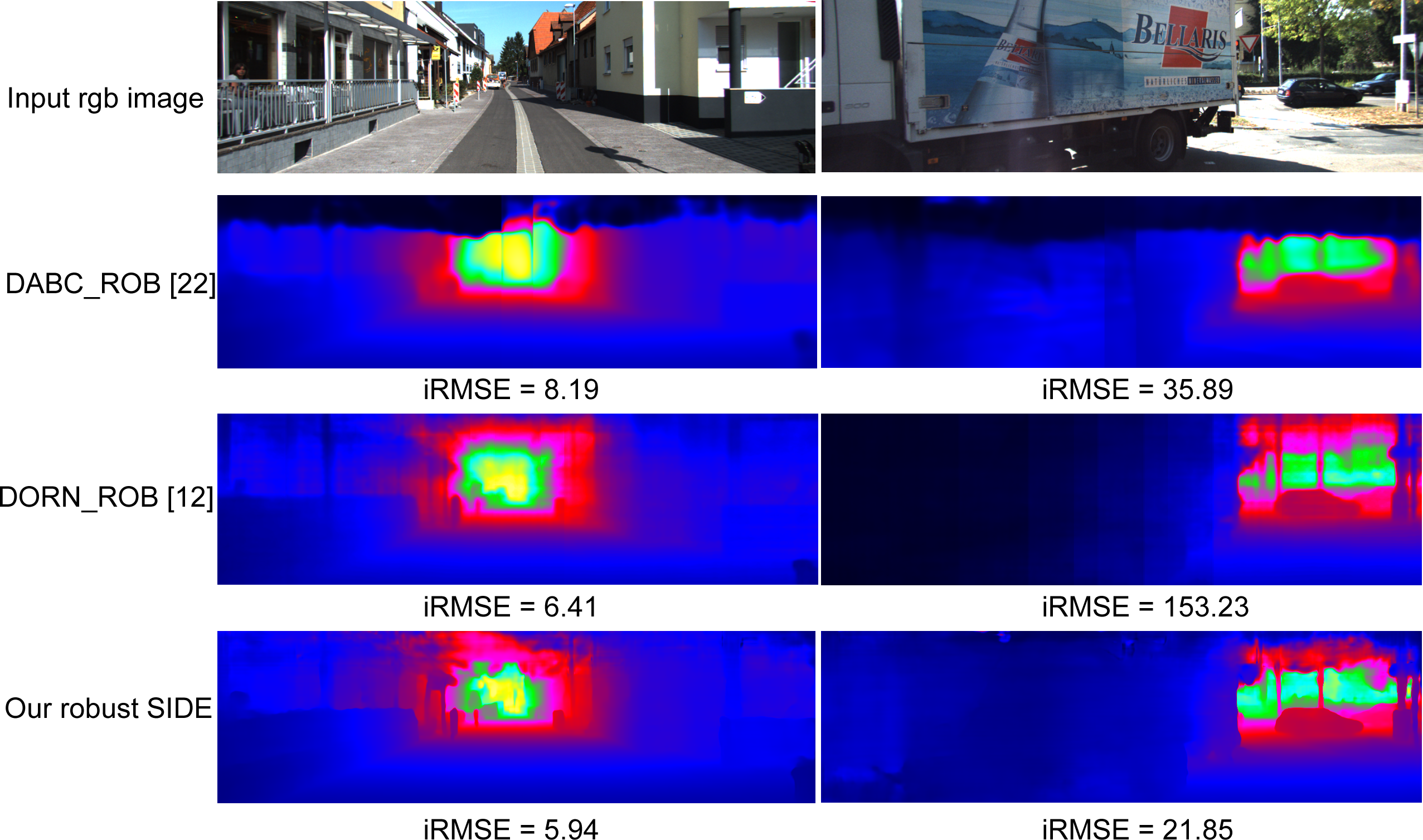}
\caption{Example outputs of our robust SIDE framework in KITTI SIDE test set, comparison to ROB submissions.}
\label{fig:KITTI}
\end{figure}

Next, we show the accuracy of our robust SIDE framework in ScanNet and KITTI, in Table \ref{table:ScanNet accuracy} and Table \ref{table:KITTI accuracy} respectively. The scene understanding is achieved by WideResNet scene classification in `Ours-sc', and by DS-SIDENet coarse depth estimation in `Ours-cde'. It can be seen that we achieve best accuracy in both the ScanNet and KITTI datasets, compared to top ROB submissions\footnote{The ScanNet leaderboard is not updated anymore. For KITTI leaderboard, please see the our DS-SIDENet-ROB submission in  \url{http://www.cvlibs.net/datasets/kitti/eval_depth.php?benchmark=depth_prediction}, ROB submissions have `ROB' in their name.}. We also notice that  the SIDE accuracy using scene classification is exactly the same  as the one using coarse depth estimation. The reason is that both of these two methods are able to achieve 100\% classification accuracy in the ROB test set when classifying ScanNet images to low depth-range, and classifying KITTI images to high depth-range. This gives us another advantage that our robust SIDE framework will not decrease the accuracy much compared to training different SIDE networks for different datasets. In Section 5.3.3, we show that the accuracy of using coarse depth estimation is sensitive to the threshold of depth-range. In realistic scenario, using scene classification will be better since it provides an independent voting into the scene types and will be less sensitive to threshold errors. We also notice that if we directly use the coarse depth as the final output (`cde only'), the accuracy is much worse. This signifies the advantage of our robust SIDE framework, compared to training one network based on dataset fusion. Due to the high-efficiency of DS-SIDENet, our robust SIDE network still takes 300-400ms on average to process the ROB testing images in single Telsa V100 GPU. Some example outputs of KITTI images are given in Fig. \ref{fig:KITTI}. 

%We show that both the scene classification network and coarse depth estimation network can achieve a 100\% accuracy for this depth-range classification on ROB testing data.
 
\begin{table}
\caption{Depth estimation accuracy in ScanNet test set, compared to the ROB submissions. `cde' means coarse depth estimation, `sc' means scene classification. }
\small
\renewcommand{\arraystretch}{1.0}
\centering
\begin{tabular}{|c|c|c|c|c|}
\hline
Method & RMSE & MAE & REL & sqREL  \\ \hline
CSWS-ROB \cite{li2018monocular} & 0.31 & 0.24 & 0.15 & 0.06 \\
DABC-ROB \cite{li2018deep} & 0.29 & 0.22 & 0.14 & 0.06 \\
DORN-ROB \cite{fu2018deep} & 0.29 & 0.22 & 0.14 & 0.06 \\ \hline
cde only & 0.366 & 0.292  & 0.191 & 0.113 \\
Ours-sc  & \textbf{0.287} & \textbf{0.219} & \textbf{0.138} & \textbf{0.057} \\ 
Ours-cde  & \textbf{0.287} & \textbf{0.219} & \textbf{0.138} & \textbf{0.057} \\ \hline

\end{tabular}
\label{table:ScanNet accuracy}
\end{table}

\begin{table}
\caption{Depth estimation accuracy in KITTI test set, compared to ROB submissions. `cde' means coarse depth estimation, `sc' means scene classification. }
\small
\renewcommand{\arraystretch}{1.0}
\centering
\begin{tabular}{|c|c|c|c|c|}
\hline
Method & iRMSE & REL (\%) & SIlog & sqREL (\%)   \\ \hline
DABC-ROB \cite{li2018deep} & 15.53 & 12.72 & 14.49 & 4.08 \\
DORN-ROB \cite{fu2018deep} & 15.96 & 10.35 & 13.53 & 3.06 \\
CSWS-ROB \cite{li2018monocular} & 16.38 & 11.84 & 14.85 & 3.48 \\\hline
cde only & 19.96 & 20.17 &  21.22 & 6.75 \\
Ours-sc & \textbf{14.61} & \textbf{10.08} & \textbf{12.88} & \textbf{2.78} \\ 
Ours-cde & \textbf{14.61} & \textbf{10.08} & \textbf{12.88} & \textbf{2.78} \\ \hline
\end{tabular}
\label{table:KITTI accuracy}
\end{table}

As summary, our DS-SIDENet is effective for single image depth estimation. Our robust SIDE framework based on DS-SIDENet is  capable of providing considerable depth estimation results for both indoor and outdoor scenarios.

\subsection{Ablation study and discussions}
\noindent{}In this section, we give ablation study of the key modules of our DS-SIDENet and robust SIDE framework.

\subsubsection{Multi-task  vs. single-task in DS-SIDENet}
First, we test DS-SIDENets with different tasks in depth estimation, which corresponds to different decoding branches. We consider three different branches, including the standard depth classification with cross-entropy loss (hard-cls), the standard depth regression (reg), and the soft classification loss (soft-cls) in DS-SIDENet. In Table \ref{table:differentheads}, we train DS-SIDENets for the above three tasks, as well as our multi-task two-branch network which combines them. It can be seen that the DS-SIDENet based on `soft-cls' loss achieves the best performance among the single-task networks. This observation is consistent to the one in disparity estimation \cite{kendall2017end}, where the cross-entropy loss performs worse compared to the one based on soft-probability. In addition, we notice that the multi-task networks clearly show better performance than the single-task networks. This can be explained as  the multi-task learning is able to improve the discriminative ability and generalization power of the encoding module. 

%So we have the conclusion that our proposed multi-task learning for depth estimation can improve the overall accuracy compared to single-task SIDE network.

\begin{table}
\caption{Accuracy comparison of DS-SIDENets with different tasks. RMSE and REL in NYU dataset are given. }
\small
\renewcommand{\arraystretch}{1.0}
\centering
\begin{tabular}{|c|c|c|c|c|}
\hline
Method & REL & RMSE & Training & Testing \\ \hline

DS-SIDENet & {0.122} & {0.518} & reg & reg \\
DS-SIDENet & {0.137} & {0.522} & hard-cls & hard-cls \\
%DS-SIDENet & {0.131} & {0.525} & soft-cls & soft-cls \\
DS-SIDENet & {0.120} & {0.511} & soft-cls & soft-cls \\ \hline
DS-SIDENet & {0.126} & {0.514} & hard-cls+reg & hard-cls \\ 
DS-SIDENet & {0.119} & {0.515} & hard-cls+reg & reg \\
DS-SIDENet & \textbf{0.113} & \textbf{0.501} & soft-cls+reg & soft-cls \\ 
DS-SIDENet & {0.117} & {0.510} & soft-cls+reg & reg \\ \hline

\end{tabular}
\label{table:differentheads}
\end{table}

\begin{figure}
\centering
\includegraphics[width=8cm]{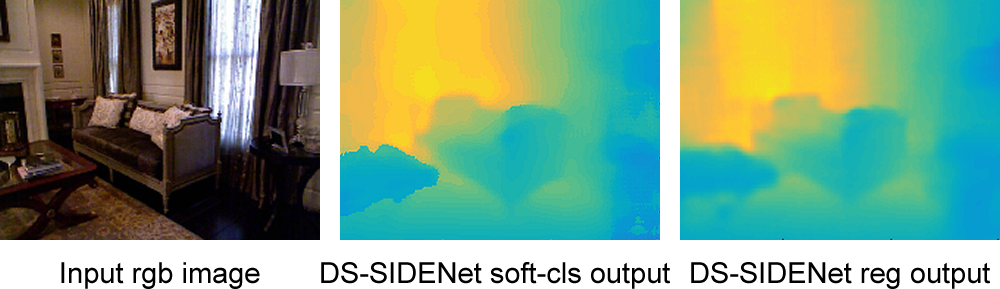}
\caption{Example output of two branches from DS-SIDENet.}
\label{fig:differenthead}
\end{figure}

In Fig. \ref{fig:differenthead}, we visualize the outputs of reg and soft-cls branches of our DS-SIDENet. It can be seen that comparing to the regression branch, the output of the classification branch is  sharper. These two branches reflect different characteristics. This motivates us to further train a fusion network \cite{ren2017image} over the outputs of these two branches. Our preliminary results show that the accuracy can be further enhanced by adding such fusion module.

\subsubsection{Using different decoding layers in DS-SIDENet}
Next, we compare the DS-SIDENets using different decoding layers. In Table \ref{table:decodingcompare}, we give the comparison between our decoding layer `upproj-con'  (row 5) and the standard `upproj' layer  (row 2), as well as with and w/o skip connections  (row 3/4). It can be seen that the DS-SIDENet with our `upproj-con' layers shows superior performance than others in NYU dataset. If we remove either the skip connection from the encoding part, or the skip connection from the RGB input, the accuracy will decrease. The reason is that the skip connection of the encoding part includes some high-level information, and the skip connection of the RGB input is similar to a pooling layer which captures global context. Integrating these information into the decoding process will be helpful.

\begin{table}
\caption{Comparison of different decoding layers in DS-SIDENet. RMSE and REL in NYU dataset are given. }
\small
\renewcommand{\arraystretch}{1.0}
\centering
\begin{tabular}{|c|c|c|c|c|}
\hline
Method & REL & RMSE & Skip-encoding & Skip-RGB  \\ \hline
DS-SIDENet & {0.126} & {0.556} & no & no \\
DS-SIDENet & {0.119} & {0.527} & yes & no \\
DS-SIDENet & {0.121} & {0.532} & no & yes \\
DS-SIDENet & \textbf{0.113} & \textbf{0.501} & yes & yes \\ \hline

\end{tabular}
\label{table:decodingcompare}
\end{table}

In Table \ref{table:KITTISHG}, we give the accuracy of the high depth-range DS-SIDENets in KITTI validation dataset, with and w/o SHG/SAM in the decoding part. It can be seen that the accuracy of DS-SIDENet with SAM decoding module is better than the one with SHG decoding module, and also better than without either SHG or SAM modules. A potential reason is that the stacked hourglass provides larger receptive field as well as keeping a relatively larger feature map size, which is more important for the high depth-range images in outdoor scenarios. With carefully designed dilated convolutions, SAM can aggregate denser features while avoiding downsampling operations in SHG. We also try adding SHG or SAM module in the low depth-range DS-SIDENet, the accuracy does not improve much. In addition, we find that it is better to add  SAM module as the third decoding block. Adding such module at the fourth or fifth block will not improve the accuracy much, but the efficiency is reduced due to larger feature map size.

\subsubsection{Scene understanding stage: scene classification vs. coarse depth estimation }
We introduced two scene understanding methods in the robust SIDE, the scene classification, and the coarse depth estimation. The advantage of using scene classification is that it does not depend on actual calculation of depth threshold, so it is more practical in real scenario. But training such network requires collecting various external training data (Places-365 dataset). In contrast, when training the coarse depth estimation, we may use the same training data as the following low depth-range and high depth-range SIDE networks. But setting the threshold $\sigma$ of depth-range is difficult. The $\sigma=5.89m$ obtained from ROB validation data can  achieve 100\% classification accuracy in the ROB testing images, but this is not realistic in real scenario. To evaluate the influence, we further do the following experiments

- R1: Keep the same coarse depth estimation, low depth-range and high depth-range DS-SIDENets trained for the optimal threshold of 5.89m, but change the threshold to 10m

- R2: Keep the same coarse depth estimation and high depth-range DS-SIDENets, but re-train low depth-range DS-SIDENet by mixing all ScanNet images with KITTI images whose maximum depth is lower than 10m. Change the threshold to 10m as well

In Table \ref{table:threshold}, we give the accuracy of the above two experiments. It can be seen that if the threshold is  set to 10m (R1), the KITTI accuracy  decreases without re-training low depth-range DS-SIDENet. The reason is that some of KITTI images will incorrectly utilize the low depth-range DS-SIDENet for depth estimation. However, the ScanNet accuracy will not change since all ScanNet images are still classfied as low depth-range images. If we re-train the low depth-range DS-SIDENet by adding some of KITTI training data (R2), although the KITTI accuracy improves, but the ScanNet accuracy decreases. The reason is that the outdoor context brought by KITTI images is totally different from the indoor context of ScanNet images. As a result, the convergence of the re-trained low depth-range DS-SIDENet will be influenced. This tells us that using coarse depth estimation for scene understanding in robust SIDE (`ours-cde') is sensitive to the threshold. In contrast, the one based on scene classification (`ours-sc') has better generalization ability in real scenario.

\begin{table}
\caption{Depth estimation accuracy of different high depth-range DS-SIDENets in KITTI validation set, with and w/o SHG/SAM module. }
\small
\renewcommand{\arraystretch}{1.0}
\centering
\begin{tabular}{|c|c|c|c|}
\hline
DS-SIDENet & SHG/SAM in decoding& iRMSE & REL   \\ \hline
+SHG & Block 3&  {8.29} & {0.075}  \\ \hline
+SAM & Block 3 & {7.83} & {0.066}  \\ \hline
+SAM & Block 4 & {7.84} & {0.066}  \\ \hline
+SAM & Block 5 &  {7.84} & {0.066}  \\ \hline
noSHG/SAM & N/A & {8.68} & {0.085}  \\ \hline
\end{tabular}
\label{table:KITTISHG}
\end{table}

\begin{table}
\caption{Accuracy of different robust SIDE methods in ScanNet  test set (RMSE/REL) and KITTI validation set (iRMSE/REL). $\sigma$ is the threshold to decide the low depth-range and high depth-range. `Re-train' means whether retraining the low depth-range DS-SIDENet by adding the KITTI images with maximum depth lower than $\sigma$. }
\small
\renewcommand{\arraystretch}{1.0}
\centering
\begin{tabular}{|c|c|c|c|c|}
\hline
Method & $\sigma$ & Re-train & ScanNet & KITTI \\ \hline
Ours-sc & N/A & No & \textbf{0.287/0.138} & \textbf{7.83/0.066}  \\ 
Ours-cde & 5.89m & No & \textbf{0.287/0.138} & \textbf{7.83/0.066}  \\ \hline
R1 & 10m & No & \textbf{0.287/0.138} &  8.44/0.072 \\ \hline
R2 & 10m & yes & 0.366/0.191 & 8.40/0.071 \\ \hline
\end{tabular}
\label{table:threshold}
\end{table}

\section{Conclusion}
\noindent{}In this paper, we proposed a two-stage framework for robust single image depth estimation. A scene understanding module is first applied to categorize the images into low depth-range and high depth-range classes.  Different SIDE networks trained specifically for these depth-ranges are utilized to obtain an accurate depth map. In addition, we developed a single image depth estimation network (DS-SIDENet) having an encoding-decoding architecture. The use of depthwise separable convolution in the encoding part ensures a better efficiency. The multi-task training in the decoding network further improves the accuracy compared to single-task network. We showed that our proposed method achieves competitive performance compared to  state-of-the-art algorithms on NYU, ScanNet, and KITTI datasets. We achieved the top rank compared to ROB 2018 submissions. 

{\small
\bibliographystyle{ieee_fullname}
\bibliography{egbib}
}

\end{document}